\documentclass{article}
\usepackage{spconf,amsmath,graphicx}

\usepackage{multirow}
\usepackage{multicol}
\usepackage{xcolor}
\usepackage{booktabs}

\title{Token-Weighted RNN-T for Learning from Flawed Data}
\name{Gil Keren, Wei Zhou, Ozlem Kalinli}
\address{Meta AI}

\begin{document}
%
\maketitle
\begin{abstract}
    ASR models are commonly trained with the cross-entropy criterion to increase the probability of a target token sequence.
    While optimizing the probability of all tokens in the target sequence is sensible, one may want to de-emphasize tokens that reflect transcription errors.
    In this work, we propose a novel token-weighted RNN-T criterion that augments the RNN-T objective with token-specific weights. The new objective is used for mitigating accuracy loss from transcriptions errors in the training data, which naturally appear in two settings: pseudo-labeling and human annotation errors. 
    Experiments results show that using our method for semi-supervised learning with pseudo-labels leads to a consistent accuracy improvement, up to 38\% relative.
    We also analyze the accuracy degradation resulting from different levels of WER in the reference transcription, and show that token-weighted RNN-T is suitable for overcoming this degradation, recovering 64\%-99\% of the accuracy loss. 
\end{abstract}

\section{Introduction}

Modern ASR systems have rapidly improved over the last years and are being deployed for a variety of applications \cite{hinton2012deep,chiu2018state,he2019streaming,DBLP:conf/interspeech/ZhengXKL0FKSM22,wu2020streaming,zhang2022bigssl,amodei2016deep}. One common ASR architecture is the Recurrent Neural Network Transducer (RNN-T) \cite{graves2012sequence}, whose training objective aims to increase the probability of the target sequence, summing across all possible alignments. RNN-T has become ubiquitous due to its compelling accuracy for a relatively small model size, and is deployed in a large number of production systems \cite{DBLP:conf/interspeech/ZhengXKL0FKSM22,wu2020streaming,he2019streaming}.

While increasing the probability of the target sequence is a natural training objective, it ignores the relative importance of the different tokens in the sequence. In certain applications, some tokens in the sequence may be more important than others, e.g., when emphasizing the importance of some in-domain words or punctuation symbols. In some other applications, some tokens' importance may need to be de-emphasized due to errors in the target reference text. 

Those reference text errors may appear in at least two scenarios. First, human annotation is known to be imperfect and the human transcription errors used in training may cause accuracy degradation. Second, Semi-Supervised Learning (SSL) through pseudo-labeling has been a common paradigm in ASR, and many large scale production systems rely heavily on this method \cite{kahn2020self,DBLP:conf/interspeech/ZhengXKL0FKSM22,lugosch2022pseudo,xu2020iterative,synnaeve2019end,DBLP:conf/interspeech/ChenWW20,DBLP:conf/interspeech/HiguchiMRH21,DBLP:conf/interspeech/LikhomanenkoXKS21}. With pseudo-labeling, the problem may be exacerbated as the initial model generating the pseudo-labels could be of a lower quality and introduce a large number of errors into the training data.  

In this work, we focus on the RNN-T training criterion and present a novel version thereof, that allows applying importance weights to the different tokens in the target sequence. For attention-based encoder-decoder models, the training objective is often simply the sum of log conditional token emission probabilities. Therefore, for those models,  applying token-weights may be straightforward. However, the RNN-T model only defines the full sequence probability and there is no standard definition of the local conditional token probability, 
In this work, we provide this definition with a detailed derivation and show that it can be obtained as the probability mass of all partial alignment paths resulting in a given token.

Furthermore, our method for computing token-specific emission probabilities can be repurposed to be token-specific confidence scores.
For SSL with pseudo-labeling, we use those token-specific confidence scores of the teacher model as token weights for our token-weighted RNN-T objective to train the student model. 
This leads to a token-weighted method for learning from pseudo-labels. 
In experiments, we show that this pseudo-labeling method can be applied iteratively to achieve up to 38\% relative WER improvement over baseline methods. This approach may be extended to CTC and encoder-decoder with attention models as well.

We conduct another experiment to simulate the setting of human annotation errors, and show our token-weighted RNN-T version can mitigate accuracy losses incurred from such errors. We use a pretrained model to emit token-level confidence scores for those annotations, and those scores are in turn used again as token weights for training a new model. Our experiments show that our method managed to recover 64\%-99\% over the accuracy loss caused by those annotation errors. 

In summary, this work's contributions include a novel token-weighted version of the RNN-T objective, which is also used as a novel method for emitting token confidence scores. Furthermore, we show that confidence-weighted RNN-T training results in considerable accuracy improvements for SSL and when human annotation errors exist in the training data.

\section{Related Work}
We discuss related work that is not otherwise mentioned. The problem of using low quality ASR transcriptions has been studied in previous works. In \cite{pratap2022star}, an extension of CTC training objective is proposed for the case of partially labeled data. This is implemented using a Weighted Finite-State Transducer (WFST) defining valid alignment paths and a framework for GPU differentiation of those WFSTs. In \cite{gao2023learning}, the authors augment the CTC objective by using a similar WFST that accounts for possible insertion, deletion or substitution errors in the reference transcriptions, and again use a GPU differentiation framework for WFSTs. This direction was also explored in \cite{cai2021w,DBLP:journals/corr/abs-2306-01031}. While both those methods aim at mitigating accuracy errors from low quality transcriptions, our work is aimed at RNN-T, simpler to implement, and is making use of token-level confidence scores as additional signal. In \cite{zhu2023alternative}, a token-level confidence score is compared to a threshold, and the CTC objective is extended to allow alternative tokens instead of those with low confidence. However, the authors required a non-trivial automatic threshold finding mechanism as part of this method's pipeline. In contrast, our work focuses on RNN-T and does not require any confidence threshold hyperparameter.

As an alternative approach, the problem of transcription errors in pseudo-labels was addressed in the literature by applying data filtering, often by using the model confidence in those transcriptions \cite{chen2023improving,kahn2020self,charlet2001confidence,vesely2017semi,zhang2014semi,wessel2004unsupervised,vesely2013semi,DBLP:conf/interspeech/ZhengXKL0FKSM22}. While those methods may reduce utterances with a large number of errors from the training set, they also inevitably remove many correctly transcribed tokens in those utterances. In addition, such methods typically require carefully tuning some threshold that determines which utterances are to be filtered out. 

Concurrently with our work, a recent work \cite{hu2024self} proposes to use token-level confidence scores for Attention-based Encoder-Decoder models \cite{chan2016listen} training. The confidence scores are obtained by aggregating attention scores and model prediction confidence. However, this method is not applicable for RNN-T or CTC models.

\section{Method} \label{sec:method}

\subsection{Recurrent Neural Network Transducers}

Let $x$ denote the acoustic input sequence and $y=y_1, ..., y_U$ denote the output token sequence from a vocabulary $V$.
We denote $y_{<u}$ as the partial sequence $y_1, ..., y_{u-1}$.
The RNN-T model has three main components, namely the encoder network, the prediction network and the joint network.
The encoder network encodes the input into high-level representations with optional subsampling to $T$ frames.
The prediction network encodes the emitted tokens and the joint network (the joiner) further combines the above two into token logits.
A final softmax normalization is applied on the joint network output for every combination of time step $t$ and output position $u$, which accounts for the blank-augmented alignment sequences.
This is denoted as $P_{t, u}(k)$, where $k \in V \cup \{\phi\}$ and $\phi$ is the blank symbol. 
The standard RNN-T objective minimizes $-\log P(y|x)$ as a marginalization over all alignment sequences, which is normally computed using a forward-backward algorithm \cite{graves2012sequence}.

\subsection{Token-Weighted RNN-T Objective}
We expand the sequence probability and rewrite it as the product of conditional token probabilities. Note that all probabilities introduced in this section are conditioned on the audio input $x$, therefore for ease of notation we omit $x$ from the equations in the rest of Section \ref{sec:method}.
\begin{align}
    -\log P(y) &= -\log \prod _{u=1}^{U} P(y_u | y_{<u}) \\ \nonumber
    &= -\sum _{i=u}^{U} \log P(y_u | y_{<u}).
\end{align}
This formulation allows us to naturally add token-specific weights $\lambda_u$ for each summand: 
\begin{equation} \label{eq:loss}
    L_w = -\sum _{u=1}^{U} \lambda_u \log P(y_u | y_{<u}).
\end{equation}
When the token weights all equal to one, the above token-weighted RNN-T objective amounts precisely to the standard RNN-T objective. Note that the derivation of Eq.~\eqref{eq:loss} is so far not specific to RNN-T but sequence prediction models in general.




\subsection{Calculating $P(y_u | y_{<u})$ in RNN-T} \label{sec:cond}


Applying Eq.~\eqref{eq:loss} directly into the RNN-T objective is not straightforward from a mathematical point of view, due to the marginalization over all alignment sequences.
The major problem is how to define and efficiently compute the conditional probability $P(y_u | y_{<u})$ in RNN-T.
A naive option would be to use the token emission probabilities as observed during beam search decoding. However, RNN-T token emission probabilities are often spread across multiple time steps, and standard beam search algorithms \cite{graves2012sequence,tripathi2019monotonic,boyer2021study} are not guaranteed to aggregate emission probabilities over all possible alignments. Therefore, using token emission probabilities from a beam search algorithm may be less accurate than precisely aggregating token emission probabilities across all possible alignments.  
Instead, in the following we provide a detailed derivation to define $P(y_u | y_{<u})$ in RNN-T, by precisely aggregating token emission probabilities across all possible alignments, which will further reveal the implementation insights of how to apply the method in practice.

\begin{figure*}[htb]
	
    \centering
    \includegraphics[width=17.5cm]{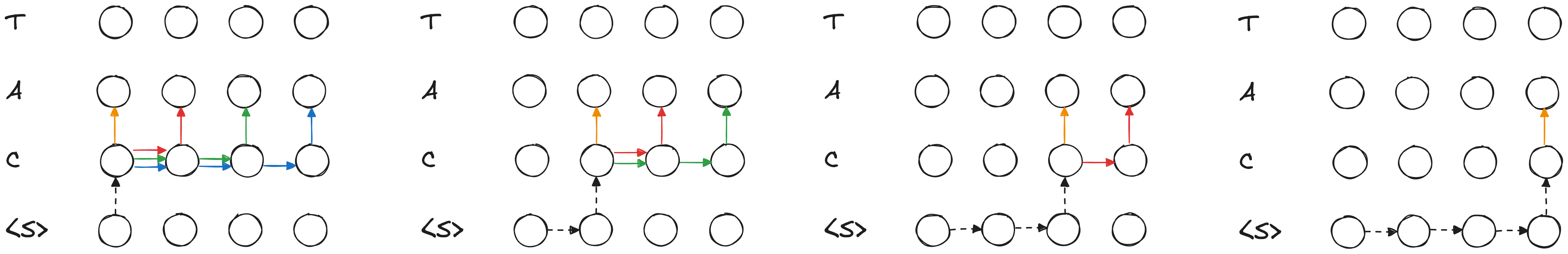}	
	\caption{All possible partial alignments from the token 'C' to the token 'A' in four audio frames. Each image corresponds to different time step where the token 'C' may have been emitted. The horizontal arrows correspond to emitting the blank symbol $\phi$. When computing $P(y_u | y_{<u}$) we sum across all the above paths.}
	\label{fig:alignments}
\end{figure*}

Informally, for modeling $P(y_u | y_{<u})$, we want to sum the probabilities of all partial alignments that start from emitting $y_{u-1}$ and end in emitting $y_u$. 
In any such partial alignment, $y_{u-1}$ is emitted in some time step $t_{u-1}$, which is followed by zero or more emissions of the blank token and subsequently emitting $y_u$ at a time step $t_u$ such that $t_u \geq t_{u-1}$. 
See Figure \ref{fig:alignments} for an illustration of the possible paths. Note that this method for calculating $P(y_u | y_{<u})$ is inspired by an RNN-T beam search algorithm presented in \cite{keren2023}, in which token emission probabilities are aggregated across a segment of frames. 
However, the beam search algorithm from \cite{keren2023} is not used in this work. 

Formally, we start with the following definition of the conditional probability:
\begin{equation} \label{eq:div}
    P(y_u | y_{<u}) = P(y_{<u+1}) /  P(y_{<u}).
\end{equation}
The right hand side of the above equation is comprised of two (unconditional) sequence probabilities. Computing the probability of entire sequences is already well defined since RNN-T's inception \cite{graves2012sequence}, and efficient algorithms using dynamic programming and low-level implementations are readily available. However, it still remains to show how to compute the above Eq.~\eqref{eq:div} efficiently and without redundancies, i.e., avoid computing two separate sequence probabilities, but instead reuse the computation of $P(y_{<u})$ when computing $P(y_{<u+1})$. Below we present a natural method of computing a sequence probability recursively, such that no additional computation is incurred compared to the standard RNN-T loss computation. 

To this end, we factorize $P(y_{<u+1})$ by the different possible time steps $t_{u-1}$ and $t_u$ in which $y_{u-1}$ and $y_u$ are emitted, respectively:
\begin{align} \label{eq:factor}
    P(y_{<{u+1}}) &= \nonumber \\
    & \sum_{t_{u-1}=1}^{t_u} P(y_{u-1}@t_{u-1} | y_{<u-1}) \nonumber \\
    & \sum _{t_u = t_{u-1}}^T  P(y_u@t_u | y_{u-1}@t_{u-1}, y_{<u-1}),
\end{align}
where $y_u@t_u$ is the subset of alignments of $y_1, ..., y_u$ in which $y_u$ is emitted in time step $t_u$.

The second part of Eq.~\eqref{eq:factor} is the probability of a single partial alignment, in which the locations $t_{u-1}$ and $t_u$ of emitting $y_{u-1}$ and $y_u$ are fixed. Therefore it is directly computed from the output of the softmax layer applied on the joint network's output, and is simply a product of the probabilities to emit the blank symbol from $t_{u-1}$ to $t_u$ and then emitting $y_u$ at $t_u$:
\begin{align}\label{eq:t1-to-t2}
    P(y_u@t_u | y_{u-1}@t_{u-1}, y_{<u-1}) &= \nonumber \\
    P_{t_u, u-1}(y_u) & \prod_{t'=t_{u-1}}^{t_u} P_{t', u-1}(\phi).
\end{align}

The first part of Eq.~\eqref{eq:factor} is computed recursively, as seen by factorizing over the locations where the previous token was emitted:
\begin{align} \label{eq:recursive}
    P(y_u@t_u | y_{<u}) &= \nonumber \\
    \sum_{t_{u-1}=1}^{t_u} & P(y_{u-1}@t_{u-1} | y_{<u-1}) \nonumber \\ 
    & P(y_u@t_u | y_{u-1}@t_{u-1}, y_{<u-1}).
\end{align}
The first probability term in Eq.~\eqref{eq:recursive} is the recursive part, while the second probability term is computed in Eq.~\eqref{eq:t1-to-t2}.
For boundary conditions, we have
\begin{align}
    P(y_0@t_0) =
    \begin{cases}
        1, & \text{if} \quad t_0 = 1 \\
        0, & \text{otherwise}.
    \end{cases}
\end{align}

Similar to \cite{zhou2021segmental-transducer}, the final blank of RNN-T model can be regarded as a special sentence end token $y_{U+1}$ that must be emitted at frame $T$.
However, this $y_{U+1}@T$ is only for completeness of the derivation, which is irrelevant for the proposed token-weighted objective.

Based on the above derivation, we showed that the conditional token probability in RNN-T can be computed via Eq.~\eqref{eq:div}, which is efficiently computed via the factorization in Eq.~\eqref{eq:factor}. The conditional token probability is then plugged into Eq.~\eqref{eq:loss} to obtain the proposed token-weighted RNN-T objective.

\subsection{Token Weights $\lambda _u$ as Teacher Confidence Scores} \label{sec:conf}

The token-weighted RNN-T can be used for a variety of use cases such as emphasizing the importance of tokens from punctuation symbols, domain-specific words, etc. In this work however, we focus on an application for mitigating reference text errors in the training set, that origin either from human annotation errors or machine-generated pseudo-labels. 
When training a student model using pseudo-labels, we propose to set the token weights $\lambda_u$ as the confidence scores from the teacher model that generates the pseudo-labels. Ideally, erroneous tokens in the training set will be (on average) associated with lower confidence scores in the teacher model, and thus, their impact on the loss and gradient of the downstream student model will be diminished.

A number of ways to obtain confidence scores appear in the literature. In this work, we extend our contributions by noting that given an already trained teacher model, 
Eq.~\eqref{eq:div} can again be applied as token confidence scores obtained by the teacher model.
More specifically, we denote $c_u := P_\text{teacher}(y_u | y{<u})$ as the token-level confidence score obtained from the teacher model for token $y_u$.
The token weights $\lambda_u$ is then defined as
\begin{align}
    \lambda_u = \frac{c_u^\alpha}{\frac{1}{U'}\sum_{u'} c_{u'}^\alpha}
\end{align}
Here $\alpha$ is a tunable scale to further control the relative importance of tokens based on their associated confidence scores. 
Additionally, to keep the the gradient absolute size relatively fixed when tuning for the best value of $\alpha$, we apply a batch-level normalization for the weights $\lambda_u$.



\section{Experiments}
We experiment with token-weighted RNN-T in two settings, pseudo-labeling and simulating human annotation errors. 

\subsection{Pseudo-Labeling} \label{sec:exp1}
Pseudo-labeling \cite{kahn2020self,DBLP:conf/interspeech/ZhengXKL0FKSM22,lugosch2022pseudo,xu2020iterative,synnaeve2019end,DBLP:conf/interspeech/HiguchiMRH21,DBLP:conf/interspeech/LikhomanenkoXKS21} is a Semi-Supervised Learning (SSL) method in which a trained model is generating hypotheses for audio segments, and those are in turn being used as reference texts for training another model. Pseudo-labeling introduces many mistakes into the training data, making it a good candidate to benefit from token-weighted RNN-T. 

\begin{table}[htb]
    \centering
    \begin{tabular}{clcc}
    \toprule
        Train & RNN-T Model & test-clean & test-other  \\
         Round & Training & WER & WER  \\ \hline
        \midrule     
        --- & 100h baseline & 10.76\% & 25.49\%  \\
        --- & 960h baseline & 2.66\% &	6.14\%   \\ \hline
        
        \midrule     
        \multirow{3}{*}{1} & Standard RNN-T Loss &  7.61\% &	17.54\%   \\
         & Utterance Weights &  7.28\%	& 17.02\% \\
         & Token-Weighted RNN-T &  \textbf{6.64\%}	& \textbf{15.55\%}   \\
        \hline
         
         \midrule     
        \multirow{3}{*}{2} & Standard RNN-T Loss & 7.34\% &	16.89\%    \\
         & Utterance Weights &  6.92\%	& 15.21\% \\ 
         & Token-Weighted RNN-T &  \textbf{5.61\%}	 & \textbf{12.32\%}   \\
         \hline
         
         \midrule     
        \multirow{3}{*}{3} & Standard RNN-T Loss &  7.15\%	& 16.4\%   \\
         & Utterance Weights &  6.78\%	& 15\% \\ 
         & Token-Weighted RNN-T &  \textbf{5.23\%}	& \textbf{11.15\%}   \\
         \hline
         
         \midrule     
        \multirow{3}{*}{4} & Standard RNN-T Loss &  7.14\%	& 16.6\%   \\
         & Utterance Weights &  6.85\%	& 14.79\% \\ 
         & Token-Weighted RNN-T &  \textbf{5.06\%}	& \textbf{10.49\%}   \\
         \hline
         
         \midrule     
        \multirow{3}{*}{5} & Standard RNN-T Loss &  6.92\% &	16.22\%   \\
          & Utterance Weights &  6.73\%	& 14.58\% \\ 
         & Token-Weighted RNN-T &  \textbf{4.94\%}	& \textbf{10.06\%}   \\
         \hline
    \bottomrule
    \end{tabular}
    \caption{SSL WERs for the Conformer RNN-T model.}
    \label{tab:conformer}
\end{table}

The Librispeech corpus \cite{panayotov2015librispeech} is used for this set of experiments. We use the labeled 100 hours training split to train our base model. The rest of the training split, 860 hours, is treated as unlabeled data that is pseudo-labeled. We perform five generations/iterations of training, where the base model generates pseudo-labels and token confidence scores for the first generation, and generation $n$ generates pseudo-labels for generation $n+1$, as described in Section \ref{sec:conf}. The confidence scores are used as token-weights as per Eq.~\eqref{eq:loss}, and we optimize over $\alpha=1, 2, 4, 6, 8$ (each value corresponds to a trained model) in each iteration using the validation set, separately for our proposed method and the baselines. For our proposed method, we find that normally large values of $alpha$ (6 or 8) result in best accuracy. Except the base model, we train with both the 100 hours labeled data and the 860 hours pseudo-labeled data, with a ratio of 1:9 in favor of the latter, which reflects the datasets' relative sizes.  


Our token-weighted RNN-T is compared against two baselines. The first is standard RNN-T. The second is a model that uses utterance-level confidence weights, which is a standard baseline for this method \cite{kahn2020self}. The utterance-level weights are the average token confidence scores across the utterance \cite{kahn2020self,DBLP:conf/interspeech/ZhengXKL0FKSM22}. Those scores are again taken to the power of $\alpha$ and normalized to have an average of one similar to manner explained in Section \ref{sec:conf}. Those utterance level weights are used when aggregating the loss across a training batch. Note that for a large $\alpha$, this method practically amounts to a data filtering method. This baseline allows us to evaluate whether there is added value in drilling down to a token-level granularity of confidence scores, or whether utterance-level confidence scores are sufficient. All methods use an identical training recipe. We also present results using the full 960 hours labeled training set, as an upper limit for our models' accuracy.


Experiments are performed using two neural network settings. A full-context RNN-T model comprising of 24 Conformer \cite{gulati2020conformer} layers, each with dimension 512, eight attention heads, a convolution kernel size of 15 and a total of 133M parameters. The second setting is a streaming RNN-T with 20 Emformer \cite{shi2021emformer,DBLP:conf/interspeech/WuWSYZ20} layers, each with attention dimension of 512, eight attention heads, a segment size of four frames and a total of 78M parameters. In both cases, the RNN-T predictor is three layers of LSTM with dimension 512. For text tokenization we use 5000 subword units \cite{kudo2018sentencepiece}.

Results for the full context and the streaming models appear in Tables \ref{tab:conformer} and \ref{tab:emformer} respectively. In both cases, throughout the training rounds, token-weighted RNN-T results in considerable WER improvements over standard RNN-T. For the full context model, the last training round results in WER of 6.92\%/16.22\%  (clean/other) for standard RNN-T and 4.94\%/10.06\% for token-weighted RNN-T, which is 28.6\%/38.0\% of relative WER improvement. For the streaming ASR model, the last training round results WER of 9.76\%/19.78\% for the standard RNN-T and 7.02\%/14.14\% for token-weighted RNN-T, a 28.1\%/28.5\% of relative WER improvement. 

Furthermore, the results show that using utterance level confidence scores is not sufficient, and there is a considerable benefit from using those with a token-level resolution. Indeed, while utterance weights improve WER over standard RNN-T by 2.7\%/10.1\% and 9.5\%/9.0\% in the last training iteration for the full context and the streaming model respectively, those number still considerably lag behind token-weighted RNN-T.


\begin{table}
    \centering
    \begin{tabular}{clcc}
    \toprule
        Train & RNN-T Model & test-clean & test-other  \\
         Round & Training & WER & WER  \\ \hline
        \midrule     
        --- & labeled 100h & 16.71\%&	34.8\%  \\
        --- & labeled 960h & 3.81\%	&9.46\%   \\ \hline
        
        \midrule     
        \multirow{3}{*}{1} & Standard RNN-T Loss &  12.39\%&	25.68\%   \\
         &  Utterance Weights &  11.92\%	&24.94\% \\
         & Token-Weighted RNN-T &  \textbf{10.73\%}	&\textbf{22.77\%}   \\ \hline
         
         \midrule     
        \multirow{3}{*}{2} & Standard RNN-T Loss & 10.8\%	&22.24\%   \\
         & Utterance Weights &  10.09\%&	21.24\% \\
         & Token-Weighted RNN-T & \textbf{8.55\%}	&\textbf{18.21\%}   \\ \hline
         
         \midrule     
        \multirow{3}{*}{3} & Standard RNN-T Loss &  10.47\%	&21.36\%   \\
         & Utterance Weights &  9.44\%&	19.35\% \\
         & Token-Weighted RNN-T &  \textbf{7.72\%}&	\textbf{15.82\%}  \\ \hline 
         
         \midrule     
        \multirow{3}{*}{4} & Standard RNN-T Loss &  9.86\%&	20.22\%   \\
         & Utterance Weights &  9.34\%	&18.47\% \\
         & Token-Weighted RNN-T &  \textbf{7.22\%}	&\textbf{14.8\%}   \\ \hline
         
         \midrule     
        \multirow{3}{*}{5} & Standard RNN-T Loss &  9.76\%&	19.78\%   \\
         & Utterance Weights &  8.83\%	&17.79\% \\
         & Token-Weighted RNN-T &  \textbf{7.02\%}	& \textbf{14.14\%}   \\ \hline
    \bottomrule
    \end{tabular}
    \caption{SSL WERs for the Streaming Emformer model.}
    \label{tab:emformer}
\end{table}

\subsection{Simulating Human Labeling Errors}
Another source of transcription errors comes from human annotation errors. The ASR community invests significant resources into collecting high quality corpora annotated manually by humans, but even those have a non-trivial amount of errors \cite{gao2023learning}. In this experiment, we simulate the scenario of learning from a dataset with human annotation errors, study those errors affect of model accuracy and the ability to mitigate those effects using token-weighted RNN-T. For simulating human annotation errors, we consider three types of common errors: repeating a given word in the reference text, omitting a word from the reference text, and substituting a word from the reference text with a similarly sounding word (based on character edit distance). We consider a number of prescribed error levels: 10\%,20\%,30\% and 40\% of the words, which result in those numbers of WER in the reference texts. For each word we decide randomly if an error should be inserted, based on the prescribed error level. If an error needs to be inserted, we randomly chose one of the three error types to corrupt the reference text with. 

The data used for this experiment is comprised of 10,000 hours of English video data publicly shared by Facebook users; all videos are completely de-identified. The data includes a variety of speakers from diverse backgrounds and characteristics. This large scale dataset allows us to evaluate our method in a setting closer to real-world production systems. The test set split contains 65 hours from the same data domain with distinct speakers. We omit further details of this data to support the double blind process.

The model used here is a streaming ASR RNN-T model with 20 Emformer layers \cite{shi2021emformer}, each with attention dimension of 320, four attention heads, a segment size of five frames and a total of 72M parameters. The RNN-T predictor is three layers of LSTM with dimension 512. The model is trained for 25 epochs. 

As in Section \ref{sec:exp1}, token-weighted RNN-T is compared against standard RNN-T and weighting utterances based on utterance-level confidence weights. All methods compared use an identical training recipe, and $\alpha$ is optimized as in Section \ref{sec:exp1}. 
The confidence scores for this data were generated as per Section \ref{sec:conf}, using a pretrained model identical the one described in this section, that was trained on a disjoint set of 10,000 hours from the same domain. 

The results appear in table \ref{tab:exp2}. As an upper bound for accuracy we ran the standard RNN-T with no data corruption, which resulted in 23.22\% WER. Next, as expected, larger data corruption ratios result in larger WER degradation for all three training objectives. However, token-weighted RNN-T resulted in smaller WER degradation from data corruption compared to other models. For example, with 20\% of WER in the reference texts, standard RNN-T's WER degrades to 31.88\%, but token-weighted RNN-T degrades only to 25.22\%, thus recovering 76.90\% of the WER degradation. Similarly, with 30\% WER in the reference texts, standard RNN-T's WER degrades to 35\%, while token-weighted RNN-T's WER only to 27.06\%, recovering 67.40\% of the WER degradation. Similar trend exists for 40\% of WER in the reference texts. Notably, for 10\% WER in the reference texts, token-weighted RNN-T recovers 98.76\% of the WER loss. As in Section \ref{sec:exp1}, using utterance-level confidence weights results in better WER compared to standard RNN-T, but is considerably outperformed by token-weighted RNN-T.

\begin{table}
    \centering
    \begin{tabular}{clcc}
    \toprule
        Data  & Model & Test & Degradation\\
        Corruption & & WER & Recovered \\ \hline
        \midrule   
        0\% & Standard RNN-T & 23.22\% & ---\\ \hline
        \midrule   
        \multirow{3}{*}{10\%} & Standard RNN-T & 27.26\% & ---\\
        & Utterance Weights & 25.89\% & 33.91\% \\
        & Token-Weighted & \textbf{23.27\%} & 98.76\% \\ \hline
        
        \midrule   
        \multirow{3}{*}{20\%} & Standard RNN-T & 31.88\% & ---\\
        & Utterance Weights & 29.07\% & 32.44\% \\
        & Token-Weighted & \textbf{25.22\%} & 76.90\% \\ \hline
         
        \midrule   
        \multirow{3}{*}{30\%} & Standard RNN-T & 35\% & ---\\
        & Utterance Weights & 31.73\% & 27.75\% \\
        & Token-Weighted & \textbf{27.06\%} & 67.40\% \\ \hline
        
        \midrule   
        \multirow{3}{*}{40\%} & Standard RNN-T & 42.11\% & ---\\
        & Utterance Weights & 35.13\% & 36.95\% \\
        & Token-Weighted & \textbf{30.11\%} & 63.52\% \\ \hline
        
    \bottomrule
    \end{tabular}
    \caption{Accuracy with different data corruption levels.}
    \label{tab:exp2}
\end{table}


\section{Conclusion}
In this work, we presented a token-weighted version of the RNN-T objective, a simple and general solution for emphasizing chosen tokens during training. Moreover, we showed its application for SSL using pseudo-labels, which resulted in consistent accuracy improvement, up to 38\% of relative WER on the Librispeech test sets. 
We also experimentally analyzed the accuracy degradation resulting from different levels of WER in the reference transcription, which are for example as high as 46\% relative WER when the reference transcription has 30\% of WER. 
Token-weighted RNN-T is shown to be suitable for overcoming those degradation, and can recover 64\%-99\% of the accuracy loss. 

In future work we plan to extend token-weighted RNN-T to its CTC counterpart, as we expect similar gains using this training objective as well. In addition, we will explore more use cases where applying token-weights during training may be sensible, and experiment with alternative methods for emitting token confidence, which may result in improved accuracy when paired with token-weighted RNN-T. Finally, we plan on integrating token-weighted RNN-T with ideas recently proposed for extending the CTC objective such as OTC \cite{gao2023learning} and STC \cite{pratap2022star}, allowing models to better overcome transcription errors.   


\bibliographystyle{IEEEbib}
\bibliography{strings,refs}

\end{document}